\documentclass[conference]{IEEEtran}
\IEEEoverridecommandlockouts
\usepackage{amsmath,amssymb,amsfonts}
\usepackage{algorithmic}
\usepackage{graphicx}
\usepackage{textcomp}
\usepackage{xcolor}

\usepackage{multirow}
\usepackage{threeparttable}
\usepackage{bm}
\usepackage{svg}
\usepackage{amsmath}
\usepackage[normalem]{ulem}
\usepackage{makecell}
\usepackage[subrefformat=parens]{subcaption}

\def\BibTeX{{\rm B\kern-.05em{\sc i\kern-.025em b}\kern-.08em
    T\kern-.1667em\lower.7ex\hbox{E}\kern-.125emX}}

\begin{document}

\title{Loss Function Considering Dead Zone \\ for Neural Networks}

\author{
\IEEEauthorblockN{Koki Inami}
\IEEEauthorblockA{\textit{Policy and Planning Sciences} \\
\textit{University of Tsukuba}\\
Tsukuba, Japan \\
s2010042@u.tsukuba.ac.jp}
\and
\IEEEauthorblockN{Koki Yamane}
\IEEEauthorblockA{\textit{Intelligent and Mechanical Interaction Systems} \\
\textit{University of Tsukuba}\\
Tsukuba, Japan \\
yamane.koki.td@alumni.tsukuba.ac.jp}
\and
\IEEEauthorblockN{Sho Sakaino}
\IEEEauthorblockA{\textit{Systems and Information Engineering} \\
\textit{University of Tsukuba}\\
Tsukuba, Japan \\
sakaino@iit.tsukuba.ac.jp}
}

\maketitle

\section*{Abstract}

    It is important to reveal the inverse dynamics of manipulators to improve control performance of model-based control.
    Neural networks (NNs) are promising techniques to represent complicated inverse dynamics while they require a large amount of motion data. However, motion data in dead zones of actuators is not suitable for training models decreasing the number of useful training data.
    In this study, based on the fact that the manipulator joint does not work irrespective of input torque in dead zones, we propose a new loss function that considers only errors of joints not in dead zones.
    The proposed method enables to increase in the amount of motion data available for training and the accuracy of the inverse dynamics computation.
    Experiments on actual equipment using a three-degree-of-freedom (DOF) manipulator showed higher accuracy than conventional methods. We also confirmed and discussed the behavior of the model of the proposed method in dead zones.

\section{Introduction}

    The dynamic characteristics of manipulators are important to improve the control performance of model-based control.
    In particular, predicting accurate input torque from the states of each joint is very useful.

    Various approaches have been taken to improve the accuracy of inverse dynamics. Equations of motion for a manipulator that make use of prior knowledge of physics have been used for the inverse dynamics computation~\cite{RefWorks:RefID:2-atkeson1986estimation}.
    This method has generality because it assumes a model that obeys the laws of physics.
    However, it is necessary to prepare a model of the control object in advance, but getting the precise model is difficult due to nonlinear factors such as static friction.
    Errors in modeling can lead to inaccurate inverse dynamics computation.

    On the other hand, the method using NNs can deal with nonlinear factors and increase the accuracy of the inverse dynamics since NNs have high representation performance~\cite{RefWorks:RefID:3-kumpati1990identification}.
    In order to improve the performance of NNs, it is important to prepare a large amount of training data to prevent overfitting~\cite{RefWorks:RefID:4-mahajan2018exploring}.

    However, in the inverse dynamics computation of manipulators, dead zones in actuators are a barrier to obtaining multiple data.
    The manipulators do not move even though the input torque changes in dead zones.
    In dead zones, the inverse dynamics cannot be defined as a function of inputs since outputs cannot be uniquely determined by inputs.

    For the reason mentioned above, the motion data in dead zones cannot be used when calculating inverse dynamics.
    Then, an NN that performs inverse dynamics computation for a multi-DOF manipulator requires motion data where all joints are not in dead zones.
    Therefore, the number of available data decreases exponentially as the number of joints increases.

    To address this problem, this study proposes a loss function to utilize information on joints that are not in dead zones.
    The proposed loss function sets the loss of the joints in dead zones to zero while the joints not in dead zones are used for learning.
    The proposed loss function enables motion data more efficiently than the conventional NNs requiring all axes not to be in dead zones.

    In this study, we conducted experiments on actual equipment and showed that the proposed method provides higher accuracy in inverse dynamics computation compared to the Newton-Euler approach (NE approach) and a conventional NN. We also compared the actual input torque with the predicted torque in each method.

\section{Related Work}

    Inverse dynamics is largely used in adaptive control~\cite{RefWorks:RefID:22-ioannou1996robust} and model predictive control~\cite{RefWorks:RefID:5-morari1999model}\cite{tsuji18:_optim_trajec_gener_model_predic}.
    In these control methods, the accuracy of the inverse dynamics computation is one of the major factors that determine the control performance.

    Also, since the accuracy of inverse dynamics computation determines the performance of reaction force observer (RFOB)~\cite{Toshiaki_Okano202120004711}\cite{umeda18:_react_force_estim_elect_actuat}, increasing the accuracy of inverse dynamics computation makes RFOB more effective.
    Furthermore, imitation learning~\cite{Saigusa}\cite{adachi18:_imitat_learn_objec_manip_based}\cite{ayumu20:_imitat_learn_based_bilat_contr}\cite{sakaino_ojies22}\cite{hayashi_ieeeaccess_2022}\cite{hayashi_icm_2021}\cite{akagawa_jia_2023} based on RFOB has been actively used in recent years, and improving the accuracy of inverse dynamics is very useful in recent control. 

    Inverse dynamics computation has been conducted either by constructing a model based on physical laws or by estimating a model from motion data.
    In modeling according to physical laws, nonlinear terms such as static friction that exist in manipulators are difficult to formulate.
    In fact, although models that take static friction into account, such as the LuGre model, have been proposed, their applicability is limited~\cite{RefWorks:RefID:6-barahanov2000sufficient}.

    Methods for estimating a model from motion data using machine learning have also been proposed.
    Machine learning methods for the inverse dynamics computation include locally weighted linear regression~\cite{RefWorks:RefID:7-schaal2002scalable}, Gaussian process regression~\cite{RefWorks:RefID:8-hewing2019cautious}\cite{RefWorks:RefID:9-nguyen-tuong2009model}, feed forward neural network (FF-NN)~\cite{lenz2015deepmpc} and recurrent neural network~\cite{hochreiter1997long}\cite{cho2014learning}\cite{RefWorks:RefID:11-rueckert2017learning}.
    Approaches that incorporate physical laws into NNs have also been proposed~\cite{RefWorks:RefID:12-lutter2019deep}\cite{RefWorks:RefID:13-sutanto2020encoding}\cite{greydanus2019hamiltonian}\cite{cranmer2020lagrangian}.
    These approaches have been said to provide better inductive bias compared to FF-NN.

    In another way to improve NNs, loss functions for evaluating errors in machine learning have been studied for a long time such as mean squared error (MSE), mean absolute error~\cite{RefWorks:RefID:16-willmott2005advantages}, and Huber loss~\cite{RefWorks:RefID:17-huber1992robust} have been proposed.

    However, as mentioned above, there are some NNs to build models considering physical laws, but few approaches consider the physics laws in the learning phase of NNs.

    Considering the above, this study utilized prior physical knowledge during the learning phase such that the reliability of the responses in dead zones is low. Then, a new loss function is proposed to consider the reliability of responses in dead zones of actuators in the learning phase of NNs.
    The experimental results showed that the accuracy of the inverse dynamics computation was drastically improved.

\section{Method}

    This study proposes a method for learning NNs that takes into account the effect of dead zones in the inverse dynamics computation of a multi-DOF manipulator.
    Dynamics computation is the calculation to obtain the angular acceleration from the angle, the angular velocity, and the torque of each joint.
    Inverse dynamics computation is the calculation to obtain the torque from the angle, the angular velocity, and the angular acceleration of each joint.
    Let the vector of joint angles, angular velocities, angular accelerations, and torque be  $\boldsymbol{\theta}$, $\boldsymbol{\dot{\theta}}$, $\boldsymbol{\ddot{\theta}}$, and $\boldsymbol{\tau}$.
    The dynamics computation and inverse dynamics computation can be expressed as (\ref{forward_dynamics}) and (\ref{inverse_dynamics}).
    
    \begin{equation}
        \label{forward_dynamics}
        f( \boldsymbol{\theta}, \boldsymbol{\dot{\theta}}, \boldsymbol{\tau} ) = \boldsymbol{\ddot{\theta}}\
    \end{equation}
    \begin{equation}
        \label{inverse_dynamics}
        f^{-1}( \boldsymbol{\theta}, \boldsymbol{\dot{\theta}}, \boldsymbol{\ddot{\theta}} ) = \boldsymbol{\tau}\
    \end{equation}

    The objective is to construct an NN that approximates $f^{-1}$.
    In the proposed method, the NN is not constrained by the model structure, hence the model structure can be changed as needed.
    We adopted a simple three-layer FF-NN to avoid model-dependent effects.

    In the following, the data of angles, angular velocities, angular accelerations, and torque at a certain time is referred to as a sample.

    In dead zones, the manipulator does not move even though torque is input.
    When dealing with the inverse problem, outputs cannot be uniquely determined by inputs, which means that the observed torque value in dead zones has a high variance.

    In the conventional method, the loss function was the root mean square of the difference between the observed torque values and the predicted torque values.
    Much of the loss was caused by the variance of the observed values.
    The model attempted to fit the predicted values to the observed values with a large variance.
    Therefore, the overall accuracy is assumed to improve if learning is performed without including samples in dead zones.
    However, if learning is performed by omitting samples in dead zones, the number of usable samples is significantly reduced.

    The proposed method improved this problem by utilizing data of joints that are not in dead zones even when some joints are in dead zones. On the contrary, in the conventional method, data of all joints is discarded even when only one joint is in dead zones. Therefore, the sample efficiency is greatly improved by the proposed method, and the accuracy of the inverse dynamics is also improved.

    The model was constructed assuming static friction force as dead zones.
    Whether the joint was in dead zones was determined by the absolute value of the angular velocity below a certain level.
    The standard deviation obtained from the data was used at a certain level to ensure a sufficient amount of data to learn inverse dynamics and to avoid the misrecognition of a stationary state as a moving state due to the influence of sensor noise.
    We defined $r(\dot{\theta}_{ij})$ so that joint $j$ in the $i$-th sample is 0 when it is in dead zones and 1 when it is not.
    The standard deviation $\sigma_j$ of the entire motion data of joint $j$ is used as the threshold value, which is expressed using the parameter $\alpha$ as (\ref{reliability}).

    \begin{equation}
        \label{reliability}
        r(\dot{\theta}_{ij}) = \left \{
            \begin{array}{l}
            0 \;\;\;(|\dot{\theta}_{ij}| \leq \alpha \sigma_j)\\
            1 \;\;\;(|\dot{\theta}_{ij}| > \alpha \sigma_j)
          \end{array}
        \right.
    \end{equation}

    In order to ignore the error of the joints in dead zones, the loss function as shown in~(\ref{new_loss_func}) is proposed with the batch size $n$.

    \begin{equation}
        \label{new_loss_func}
        L = \frac{1}{n}\displaystyle\sum_{i=1}^{n} (\frac{1}{3} \displaystyle\sum_{j=0}^{2}r(\dot{\theta}_{ij}) (\tau_{ij} - \hat{\tau}_{ij} )^2)
    \end{equation}

    \begin{figure*}[t]
        \begin{tabular}{ccc}
            \begin{minipage}[t]{0.3\linewidth}
                \centering
                \includegraphics[height=3.6cm]{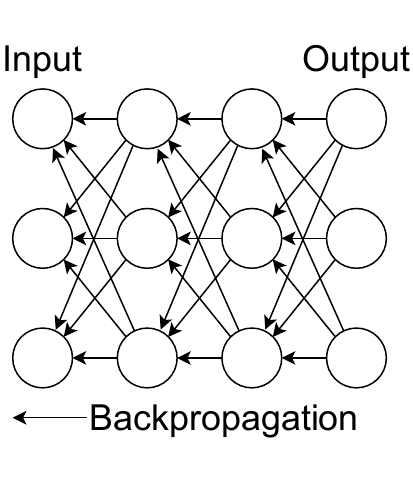}
                \vspace{\baselineskip}
                \subcaption{Conventional Method}
            \end{minipage} &
            \begin{minipage}[t]{0.6\linewidth}
                \begin{minipage}[t]{0.49\linewidth}
                    \centering
                    \includegraphics[height=3.6cm]{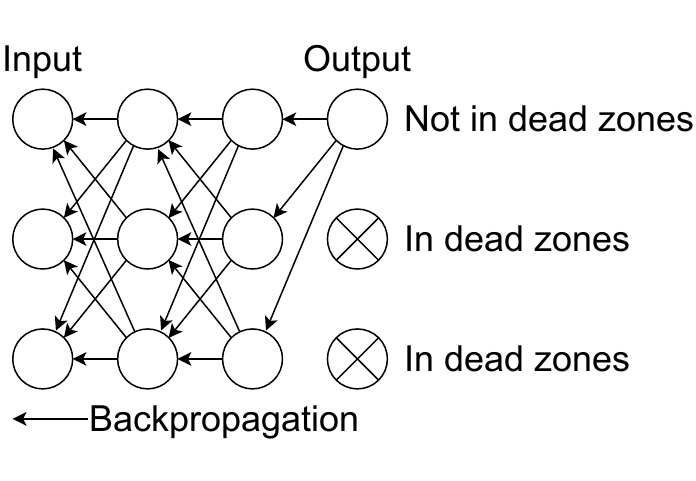}
                    \subcaption*{Some joints are in dead zones.}
                \end{minipage} 
                \begin{minipage}[t]{0.49\linewidth}
                    \centering
                    \includegraphics[height=3.6cm]{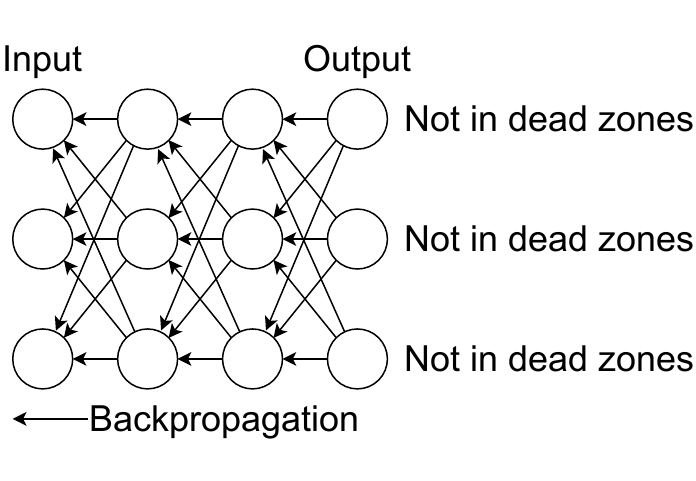}
                    \centering
                    \subcaption*{All joints are not in dead zones.}
                \end{minipage}
                \subcaption{Proposed method}
            \end{minipage}
        \end{tabular}    
        \centering
        \caption{Backpropagation in the conventional method and the proposed method}
        \label{fig:method}
    \end{figure*}

    In the proposed loss function, the square of the difference in dead zones is excluded from the loss.
    The $r(\dot{\theta}_{ij})$ is determined for each sample in the conventional method, while it is determined for each joint of the sample in the proposed method.
    The backpropagation in the proposed method is shown in Fig.~\ref{fig:method}.
    As it can be found in the left and right figures in Fig.~\ref{fig:method}, the proposed method is identical to the conventional method if there is no dead zone, while the proposed method can update the model parameters even when some joints are in dead zones (the middle figure of Fig.~\ref{fig:method}).
    Although the torque is predicted for the joints in dead zones, the error is excluded from the loss, so that it is ignored in the backpropagation during NN training and does not affect the update of the NN parameters.
    The proposed method extends the idea of dropout~\cite{RefWorks:RefID:23-srivastava2014dropout:} to the output layer and uses the physical background as a condition for making nodes inactive.
    
\section{Experimental Design}

    \begin{figure}[!t]
        \centering
            \includegraphics[width=0.65\linewidth]{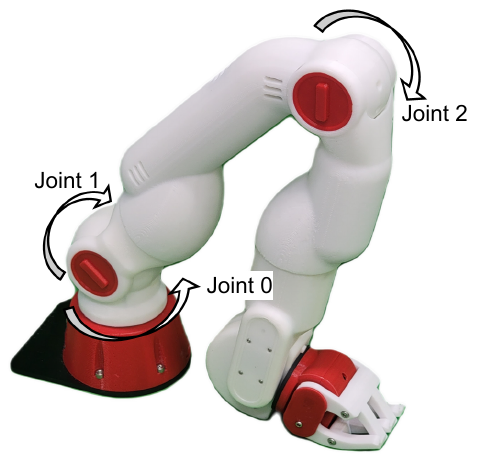}
            \caption{CRANE-X7 (RT)}
            \label{fig:CRANE-X7}
    \end{figure}

    Fig.~\ref{fig:CRANE-X7} illustrates a manipulator, called CRANE-X7,  used in the experiments. CRANE-X7 consisted of an arm with 7 DOF and a gripper with 1 DOF. In this study, the robot is used as a 3-DOF manipulator, with other joints constrained by a positional proportional-derivative (PD) controller.
    In this study, the first, second, and fourth joints of the robot are assigned numbers 0, 1, and 2, respectively.

    The control signal that determined the motor torque was obtained by using 4-channel bilateral control~\cite{Saigusa}\cite{RefWorks:RefID:20-sakaino2011multi-dof}\cite{Kosuke_Shikata202222004614}\cite{4ch}\cite{sakaino09:_obliq_coord_contr_advan_motion_contr}\cite{sakaino10:_realiz_advan_hybrid_contr_obliq_coord_contr}.
    Since 4-channel bilateral control is a remote control technology in which a human operates the leader robot, it is possible to easily obtain the data in ranges of angles, angular velocities, and angular accelerations that are close to practical.

\begin{figure}[!t]
        \centering
        \includegraphics[width=0.9\linewidth]{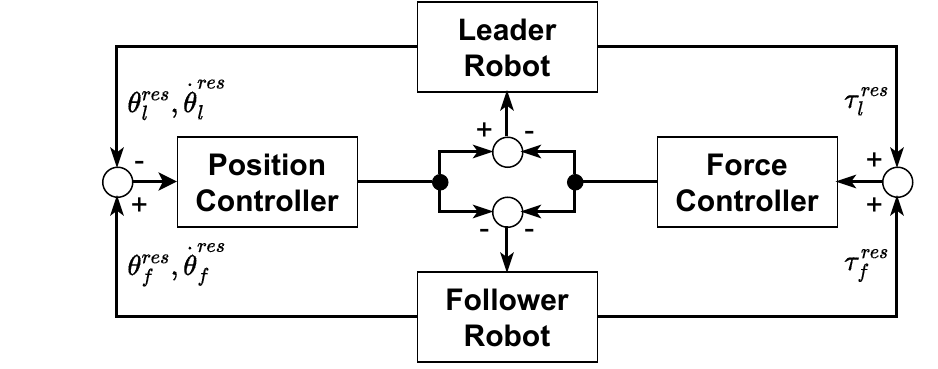}
            \caption{Block diagram of 4-channel bilateral control}
            \label{block}
    \end{figure}

    The block diagram of the 4-channel bilateral control is shown in Fig.~\ref{block}. $\boldsymbol{\theta^{res}_{l}}$ and $\boldsymbol{\theta^{res}_{f}}$ are the angles of the leader robot and the follower robot. $\boldsymbol{\dot{\theta^{res}_{l}}}$ and $\boldsymbol{\dot{\theta^{res}_{f}}}$ are the angular velocities of the leader robot and the follower robot. $\boldsymbol{\tau^{res}_{l}}$ and $\boldsymbol{\tau^{res}_{f}}$ are the reaction forces of the leader robot and the follower robot. 
    
\begin{figure}[!t]
        \centering
            \includegraphics[width=0.9\linewidth]{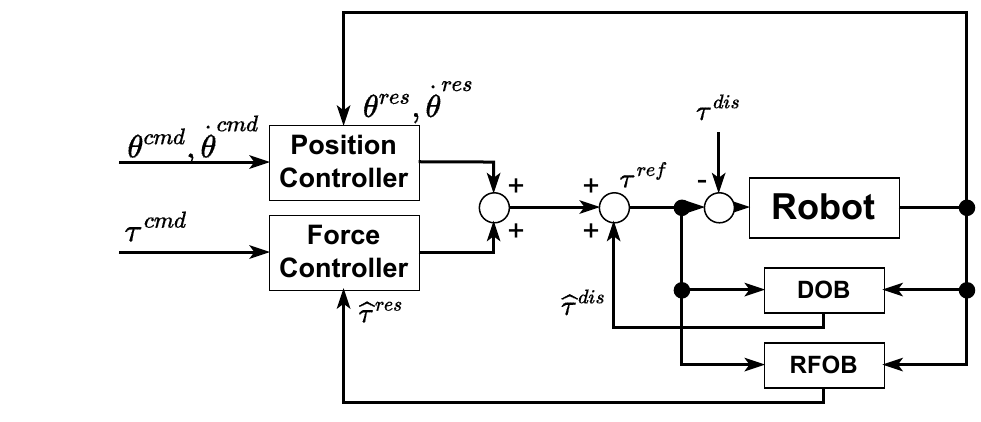}
            \caption{Block diagram of controller}
            \label{controller}
    \end{figure}

    The block diagram of the controller is shown in Fig.~\ref{controller}. The controller uses a disturbance observer (DOB)~\cite{DoB35}\cite{DOB} and RFOB to estimate external and reaction forces. The parameters of DOB and RFOB were determined in advance. $\boldsymbol{\theta^{cmd}}$, $\boldsymbol{\dot{\theta^{cmd}}}$, and $\boldsymbol{\tau^{cmd}}$ are the commands of all the angles, the angular velocities, and the torque, respectively. $\boldsymbol{\theta^{res}}$ and $\boldsymbol{\dot{\theta^{res}}}$ are the response of all the angles, the angular velocities. 
    $\boldsymbol{\tau^{dis}}$ is the disturbance and 
    $\boldsymbol{\hat{\tau}^{dis}}$ is the disturbance estimated by DOB. $\boldsymbol{\hat{\tau}^{res}}$ is the reaction force estimated by RFOB.

    \begin{figure*}[t]
    \begin{tabular}{ccc}
      \begin{minipage}[t]{0.33\linewidth}
        \centering
          \includegraphics[keepaspectratio,width=\linewidth]{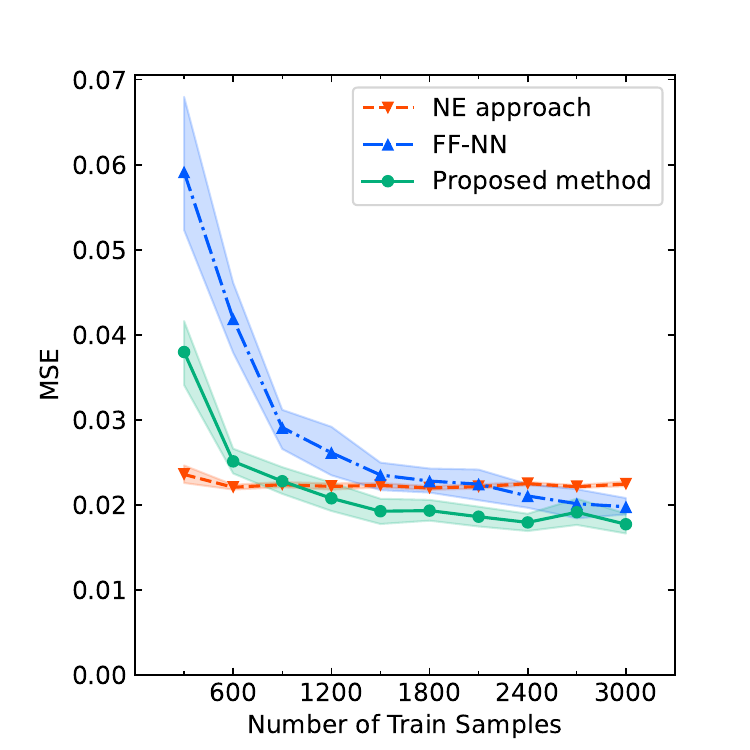}
          \subcaption{joint0}
      \end{minipage}
    
      \begin{minipage}[t]{0.33\linewidth}
        \centering
          \includegraphics[keepaspectratio,width=\linewidth]{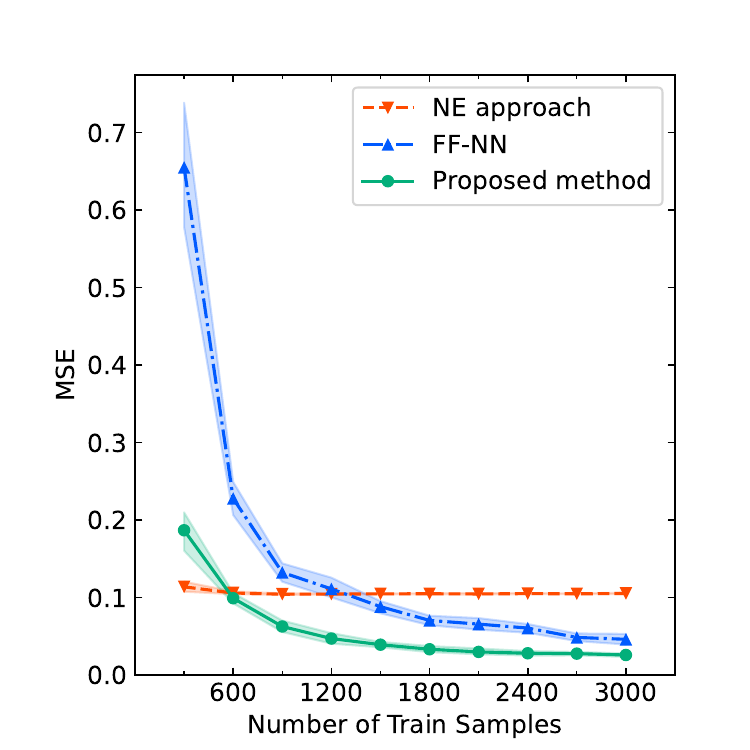}
          \subcaption{joint1}
      \end{minipage}
    
      \begin{minipage}[t]{0.33\linewidth}
        \centering
          \includegraphics[keepaspectratio,width=\linewidth]{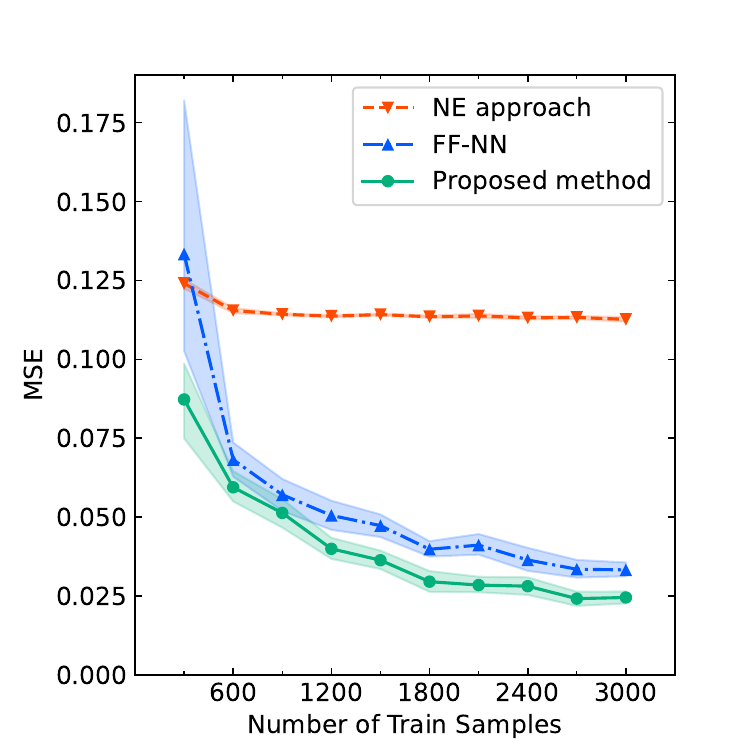}
          \subcaption{joint2}
      \end{minipage}    
    \end{tabular}
    \caption{The average MSE and 95 percent confidence interval on the test data obtained by 10 trials in which the specified number of data are randomly selected from 6000 samples of the training data are shown.}
    \label{fig:result}
    \end{figure*}
    
    By moving a leader robot appropriately and changing the command value of the follower robot, a follower robot was able to operate within a range that did not hit external objects, and motion data was obtained.
    Only the follower robot data was used to learn the inverse dynamics.

    Sampling was performed at 500 Hz for 180 seconds.
    The angular velocities and angular accelerations were derived by pseudo-differentiation of the angles obtained from the sensors.
    To compensate for the phase delay caused by the pseudo-differentiation, a low pass filter was applied once to the angular velocities and twice to the angles and torque.
    The cutoff frequency of the low pass filter was set at 25 rad/s.
    The data was down-sampled at 50 Hz and was divided into training data, validation data, and test data.

    Six thousand samples were prepared for the training data.
    To evaluate the performance based on the number of training data in increments of 300 samples from 300 samples to 3000 samples, the specified number of samples was randomly extracted from 6000 samples and used for model learning.
    Considering the variance due to sample selection, the results were evaluated 10 times for the specified sample size.
    
    For the validation and test data, 500 samples each that were not in dead zones were prepared.

    In (\ref{reliability}), $\alpha=0.1$ was used to determine dead zones.
    
    A three-layer FF-NN was adopted, and the number of nodes in the hidden layer was set to 64.
    ReLU~\cite{RefWorks:RefID:18-nair2010rectified} was used for the activation function and Adam~\cite{RefWorks:RefID:19-kingma2014adam:} was used for the optimization algorithm.
    The learning rate was set to 0.01.
    
    As control experiments, an inverse dynamics computation based on the NE approach using the recursive least square method~\cite{RefWorks:RefID:21-featherstone2014rigid} and the FF-NN with the same model structure as the proposed method were trained by MSE.
    Both methods use only samples not in dead zones for training.
    In the NE approach, the length of the links, the distance to the center of gravity of the links, the weight of the links, and the moments of inertia of the links in the Newton-Euler equations were estimated by using the recursive least squares method.

    In all methods, each training was conducted for 100 epochs and the models were evaluated using MSE with validation data.
    The MSE on the test data of the model with the epoch that showed the best MSE on the validation data was used as the final evaluation metric.
    This means that we conducted an early-stopping~\cite{ying2019overview} when the model came to an overfitting.

    When all samples were not affected by dead zones, $r(\dot{\theta}_{ij})$ was equal to 1 for all joints, and the designed loss function was the same as MSE.
    Therefore, it is safe to use MSE on the validation and the test data as an evaluation metric.

    In addition, the predicted torque obtained by the models of each method trained using 6000 samples were compared with the actual input torque. For predicting torque, we used continuous 3-second data that was not used for training.

\section{Result}

    \begin{figure*}[t]
    \begin{tabular}{ccc}
      \begin{minipage}[t]{0.33\linewidth}
        \centering
          \includegraphics[keepaspectratio,width=\linewidth]{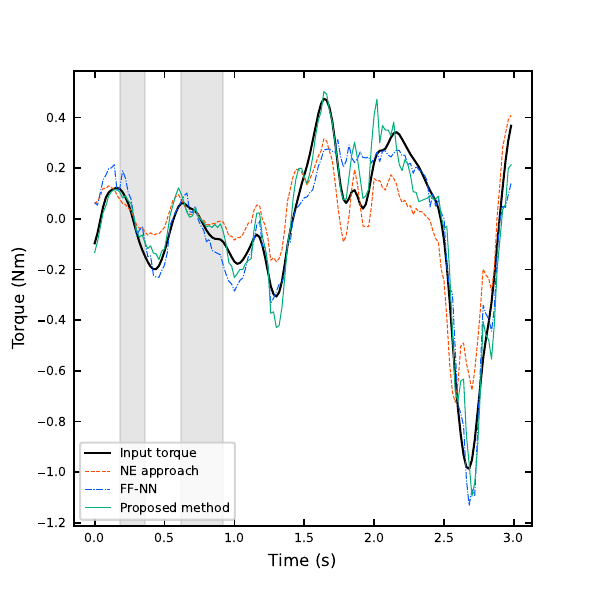}
          \subcaption{joint0}
      \end{minipage}
    
      \begin{minipage}[t]{0.33\linewidth}
        \centering
          \includegraphics[keepaspectratio,width=\linewidth]{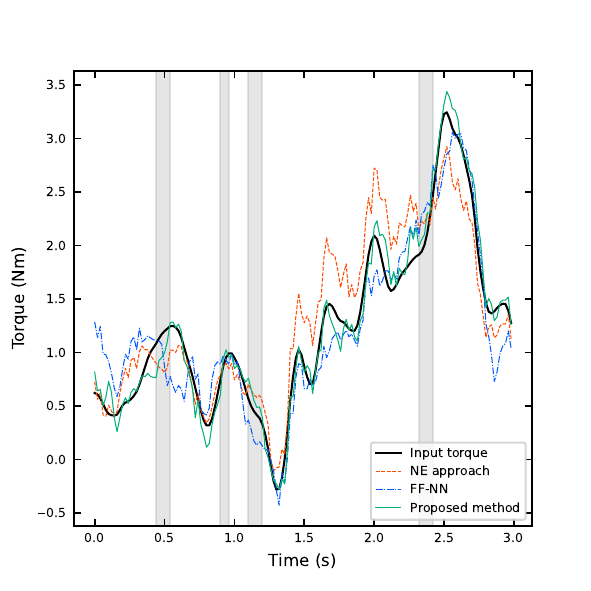}
          \subcaption{joint1}
      \end{minipage}
    
      \begin{minipage}[t]{0.33\linewidth}
        \centering
          \includegraphics[keepaspectratio,width=\linewidth]{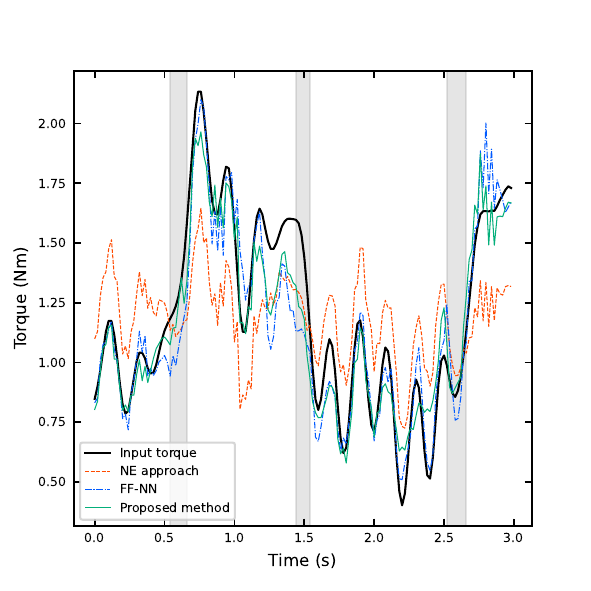}
          \subcaption{joint2}
      \end{minipage}    
    \end{tabular}
    \caption{The input torque and predicted torque are shown. The times at which each joint is considered to be affected by dead zones are shown in gray.}
    \label{fig:predicted_torque}
    \end{figure*}

    In the training data, the percentage of joints unaffected by dead zones was approximately 70 percent in one joint.
    Three joints were independently affected by dead zones, and the number of samples in which all the joints were not affected by dead zones was about 34 percent according to $0.7^{3}=0.343$.
        
    The experimental results using a real 3-DOF manipulator are shown in Fig.~\ref{fig:result}.
    The mean and 95 percent confidence interval of the MSE obtained by sampling 10 times are described in the figure.
    
    The NE approach showed similar MSE in all joints, nearly independent of the number of samples.
    Because the number of the parameters of the NE approach was small, it required very few numbers of training data.

    In all joints, the FF-NN and the proposed method showed a trend of increasing accuracy as the number of samples increased.

    In joint 0 and joint 1, the NE approach showed a smaller MSE only when the number of samples was small.
    However, when the number of samples was large, the FF-NN and the proposed method showed lower MSE.
    In joint 2, the FF-NN and the proposed method showed lower MSE than the NE approach.
    In addition, the proposed method was generally superior to the FF-NN in MSE.

    In joint 0 and joint 1, since the inertia terms vary greatly depending on the posture of the subsequent joints, the model using NN seems to require a certain number of samples or more to achieve accuracy.

    By contrast, in joint 2, the inertia term does not vary under the assumption that the manipulator is a rigid body, which seems to improve the performance of the NN with a smaller number of samples.

    One of the factors that improved the accuracy was the increased amount of data used for learning owing to the new loss function.

    While the conventional method can use only about 34 percent of the data, the proposed method can use about 70 percent of the data.
    In fact, in the joint 0 and 1, the FF-NN with twice as many samples shows the same level of MSE as the proposed method.
    This trend was not observed in joint 2 since the convergence of the FF-NN accuracy was faster.
    
    The predicted torque is shown in Fig.~\ref{fig:predicted_torque}.
    Gray bands are added at times when each joint is considered to be affected by dead zones.
    
    For all joints, the proposed method seems to best fit the input torque.
    This is similar to the trend shown by MSE.

    What is noteworthy is that in the gray bands, the proposed method appears to fit the NE approach.
    The gray bands correspond to the interpolation of the training data in the proposed method, and the proposed method seems to be trying to model a situation without dead zones. Therefore, it seems that the model based on the proposed method is able to be used even in a static state.
    
\section{Conclusion}

    To address the problem of significantly reduced available data due to dead zones in training NNs that approximate inverse dynamics computation, we proposed the loss function that ignored errors in dead zones.
    This improved the accuracy of inverse dynamics computation by increasing sampling efficiency.
    Compared to the NE approach, the proposed method showed higher accuracy when the number of training samples was sufficient.
    The accuracy of the proposed method tended to increase as the number of training samples increased as with FF-NN and was generally better than the accuracy of FF-NN.
    The model of the proposed method trained without using data in dead zones showed a similar behavior as the NE approach in dead zones.
    
    The 3-DOF manipulator was used in this study.
    As the DOF of the manipulator increases, the probability that all joints are not in dead zones decreases.
    Therefore, the proposed method is expected to be more effective in 6 or 7-DOF manipulators commonly used industrially.
    
    The proposed method can be applied to inverse dynamics computation using previously proposed machine learning models. Future work should demonstrate its effectiveness in various model structures.

\section*{ACKNOWLEDGMENT}
This study is supported by the Japan Society for the Promotion of Science through a Grant-in-Aid for Scientific Research (B) under Grant 21H01347.
    
\bibliographystyle{unsrt}
\bibliography{export}

\end{document}